\documentclass[
]{ceurart}

\usepackage{listings}
\usepackage{tcolorbox}
\begin{document}


\conference{}

\title{Learnable Assessment Skills for LLM-based Automated Scoring: Rubric Construction via Iterative Optimization}


\author[1]{Yun Wang}[%
orcid=0009-0004-6611-0752,
]
\address[1]{School of Computing, University of Georgia, Athens, GA, USA}

\author[2]{Xin Xia}[%
orcid=0009-0009-1717-8511,
]
\address[2]{AI4STEM Education Center, University of Georgia, Athens, GA, USA}

\author[1]{Xuansheng Wu}[%
orcid=0000-0002-7816-7658,
]

\author[2]{Xiaoming Zhai}[%
orcid=0000-0003-4519-1931,
email=Xiaoming.Zhai@uga.edu,
]
\fnmark[1]

\author[3]{Ninghao Liu}[%
orcid=0000-0002-9170-2424,
email=ninghliu@polyu.edu.hk,
]
\fnmark[1]
\address[3]{The Hong Kong Polytechnic University, Hong Kong, China}

\fntext[1]{Corresponding author.}

\begin{abstract}
  LLM-based automated scoring approaches near-human performance, but scaling to new tasks remains bottlenecked by the per-item human configuration of upstream stages such as rubric construction. Human experts bypass this bottleneck through evaluation heuristics developed over extensive practice. We ask whether LLMs can learn similar heuristics directly from scoring experience, and formalize this as the concept of \emph{assessment skills}: item-independent natural-language procedural knowledge that guides LLMs through specific stages of the scoring workflow. Focusing on rubric construction as a first instantiation, we propose an iterative framework that decomposes a skill into a fixed scaffold and learnable item-agnostic rules, refining the rules through LLM-driven diagnosis of scoring errors and validation-gated selection. The framework requires no expert-written rubric. On all ten ASAP-SAS items, optimized skills substantially improve LLM-based scoring and frequently surpass the dataset-provided expert rubric. Cross-item transfer experiments further reveal that learned skills capture both generalizable and item-specific patterns.
\end{abstract}


\maketitle

\section{Introduction}

Large language models (LLMs) can score open-ended student responses with agreement close to human raters on many benchmark tasks~\cite{pack2024large,impey2025using}. 
However, these systems do not scale well to \textit{new assessment items}. For each new item, human experts must still configure several upstream stages, including task interpretation, rubric design, and evidence criteria~\cite{tang2023agibench}.
This per-item setup, rather than the scoring model itself, is the main barrier to large-scale deployment.
Human experts follow the same workflow, but they rarely start from scratch. They rely on prior experience and structured knowledge, such as rubrics and learning progressions, to interpret student responses and transfer scoring criteria across tasks. This reduces the effort required to score new items.

Inspired by this, we ask whether LLMs can acquire and reuse similar knowledge for automated assessment. We define this knowledge as \textbf{assessment skills}: reusable, natural-language procedural knowledge that guides an LLM through a specific stage of the scoring workflow. Examples include deciding how to construct a rubric, how to identify evidence in a response, and how to generate feedback after scoring. We represent these skills in natural language so that they remain interpretable, inspectable, and reusable without model re-training.

Among various assessment skills, this paper focuses on \textbf{rubric construction} as a pilot case.
Rubrics define the scoring criteria and score boundaries for an assessment item, and their quality directly affects all downstream scoring decisions~\cite{yamamoto2017automated}. 
However, most existing rubrics are designed primarily for human raters. They often rely on qualitative expressions such as “partially correct” or “demonstrates basic understanding,” which assume commonsense reasoning of humans and flexible human judgment. LLMs interpret these descriptions less consistently and often produce systematic scoring errors when using the same rubrics~\cite{xue2026consistency}.
Our experiments on all ten ASAP-SAS items show this mismatch clearly: on four items, using expert-authored rubrics for LLMs actually lowers scoring performance compared to providing no rubric at all.
Meanwhile, adapting human-oriented rubrics for LLMs requires substantial manual effort and does not scale~\cite{xia2026using}. For entirely new tasks, no rubrics may be available at all.
These limitations make automatic rubric construction skills both necessary and practically important.


To address this problem, we propose an iterative optimization framework that learns rubric construction skills directly from scoring practice, without requiring any expert-written rubric as input. 
We decompose a skill into: (1) a fixed scaffold $s_0$ shared across items; (2) a learnable set of item-agnostic rules $\Delta$.
The scaffold provides the basic procedure for rubric construction, while $\Delta$ captures refinements learned from scoring errors. A rubric is item-specific, but the skill that generates it is reusable across items. This allows optimization to operate at the strategy level rather than on a single rubric.
At each iteration, the current skill generates a rubric for the target item, scores a batch of student responses, and compares the predictions with human labels. A diagnoser LLM analyzes the resulting errors and proposes an updated $\Delta$. 
This loop converts discrepancies between model and human scores into the supervision signal, allowing the framework to bootstrap rubric construction skill from human scores.

Our main contributions are as follows:
\begin{enumerate}
    \item We introduce the concept of \textit{assessment skills}, a form of learnable natural-language procedural knowledge that guides LLMs through specific stages of the scoring workflow. This formulation reduces the per-item human configuration that current automated scoring systems depend on.
    \item We instantiate this concept for rubric construction and propose an iterative framework that learns rubric construction skills from scoring practice alone, without expert-written rubrics.
    \item We evaluate the framework on the ASAP-SAS dataset and show that optimized skills improve LLM-based scoring performance across most items, often outperforming expert-provided rubrics. Further analysis demonstrates that the learned skills exhibit partial cross-item transfer, indicating that they capture both generalizable and item-specific scoring patterns.
\end{enumerate}

\section{Related Work}
\label{sec:related}
 
\paragraph{LLM-based Automated Scoring.}
Recent work explores using LLMs to support the full grading pipeline, including rubric design, scoring, and post-grading review~\cite{xie2025grade}.
Among the stages of this pipeline, rubric quality has been shown to be particularly critical for scoring reliability~\cite{wang2026autoscore}.
Tang et al.~\cite{tang2026designing} show that fine-grained, checklist-based rubrics yield higher LLM--human agreement than holistic ones in physics exams, highlighting the importance of rubric design for LLM scorers.
Along this line, Chu et al.~\cite{chu2024llm} optimize grading guidelines through self-reflection on scoring errors, and in subsequent work~\cite{chu2026confusion} further refine rubrics by decomposing misclassification patterns with a confusion matrix and applying targeted repairs to dominant error modes.
Wei et al.~\cite{weiqurl} use question-specific rubrics as verifiable reward signals for reinforcement learning in open-ended question answering.
Our work differs from these approaches in what is being optimized. Rather than optimizing a rubric or guideline for a specific item, we optimize the skill that generates them, keeping the skill itself item-agnostic.
 
\paragraph{Self-Evolving LLMs.}
Our optimization process, in which an LLM iteratively learns from its own scoring errors, connects to a growing body of work on self-evolving LLMs.
Reflexion~\cite{shinn2023reflexion} enables agents to learn from failures via verbal self-reflection, storing the reflections in an episodic memory and replaying them in subsequent trials.
ExpeL~\cite{zhao2024expel} extracts reusable natural-language insights from accumulated agent experiences and retrieves relevant ones at inference time.
Wang et al.~\cite{wang2025inducing} take a similar store-and-recall approach, but in the form of programmatic skills induced from agent trajectories.
GEPA~\cite{agrawal2025gepa} and Feedback Descent~\cite{lee2025feedback} take a different route, directly optimizing the prompt or text artifact that is then applied to the task.
Across these methods, the learned product is either applied directly or retrieved at inference time.
Our optimization target is instead a skill that, given an item, generates a task-specific rubric, which then guides downstream scoring.

\begin{figure}[t]
\includegraphics[width=\textwidth]{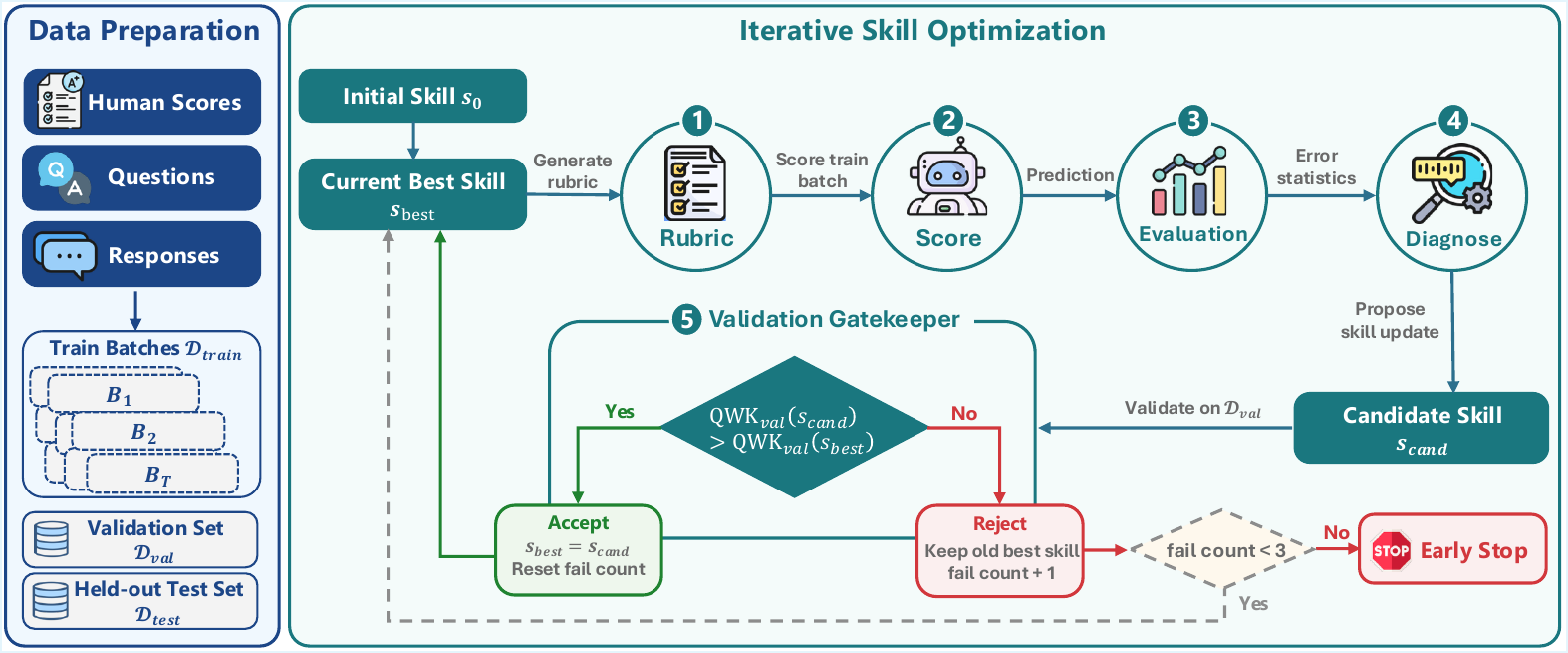}
\caption{
Overview of the iterative skill optimization framework. The system's input includes human-scored student responses, an assessment item, and a human-authored initial skill $s_0$. At each iteration, the current best skill generates a rubric (Step 1), which is used to score a training batch (Step 2). Predicted scores are compared against human scores to produce error statistics (Step 3), and a diagnoser identifies systematic error patterns and proposes an updated candidate skill (Step 4). A validation gatekeeper (Step 5) accepts the candidate only if it improves QWK on the validation set. Otherwise, the current best skill is retained. The process terminates via early stopping after three consecutive rejections or when all training batches have been used.
}
\label{architecture}
\end{figure}

\section{Methodology}

\subsection{Problem Formulation}
We consider the task of automated scoring, where the goal is to assign a score $\hat{y}$ to a student response $x$ for a given assessment item $q$. Each item has a discrete score range $\{0,1,\ldots,K\}$ and a set of human-scored examples $\mathcal{D}=\{(x_i,y_i)\}_{i=1}^m$, where $y_i$ denotes the human-assigned score for response $x_i$.
For each assessment item $q$, scoring is typically guided by a rubric $R$, which defines the scoring criteria for $q$ and how those criteria map to scores. Given a rubric $R$, an LLM scorer assigns a score by checking the response against the rubric's criteria, written as $\hat{y}_i = \mathrm{LLM}_{\mathrm{score}}(R,q,x_i)$.

In this paper, we study the more realistic setting in which no effective LLM-ready rubric $R$ can be assumed for a \textit{new item} $q$. To enable scoring under this setting, we introduce an \textbf{assessment skill} $s$, a natural-language instruction that tells an LLM \emph{how to construct rubrics for new items} rather than directly specifying any particular rubric. Given an item $q$, the skill $s$ guides the LLM to generate an appropriate rubric $R$, written as $R = \mathrm{LLM}_{\mathrm{gen}}(s,q)$, which is then used for scoring.

We measure the effectiveness of a skill by the degree of agreement between human scores and the LLM-predicted scores produced under the rubric it generates.
Formally, given a skill $s$, the predicted score for response $x_i$ is $\hat{y}_i^{(s)} = \mathrm{LLM}_{\mathrm{score}}(\mathrm{LLM}_{\mathrm{gen}}(s, q), q, x_i)$.
We measure agreement using Quadratic Weighted Kappa (QWK), a standard metric for ordinal scoring agreement.
The optimization objective is then to find the skill $s^*$ that maximizes QWK on a held-out validation set $\mathcal{D}_{\mathrm{val}}$:
\[
s^* = \arg\max_s \mathrm{QWK}(\{\hat{y}_i^{(s)}\}, \{y_i\}), 
\qquad (x_i,y_i) \in \mathcal{D}_{\mathrm{val}} .
\]

The objective above defines the optimization criterion, but not what parts of the skill should be allowed to change during optimization.
If the optimizer is allowed to rewrite the entire skill, updates may overwrite the initial guidance that encodes human prior knowledge, such as assessment design principles, rather than only adding reusable refinements to it. 
We therefore \textbf{decompose the skill} into two components $s = s_0 \oplus \Delta$.
Here, $s_0$ is a human-authored scaffold shared across all items, and $\Delta$ is a learned augmentation that extends $s_0$ with rubric construction rules acquired through optimization. 
The form of $s_0$ is flexible, with weaker scaffolds leaving more freedom for $\Delta$, while stronger scaffolds encode more detailed workflows or domain-specific frameworks such as learning progressions.

\subsection{Iterative Skill Optimization}
\label{sec:archi}

Building on the decomposition introduced above, and with $s_0$ fixed by design, the search for $s^*$ amounts to finding an effective $\Delta$.
Because $\Delta$ is an open-ended natural-language refinement rather than a differentiable parameter, we optimize it iteratively, using scoring errors as the supervision signal.
The rationale is that disagreements between predicted and human scores reveal where the generated rubric failed to guide scoring, and these failures provide evidence for how $\Delta$ should be revised.

To support this error-driven refinement process, the dataset $\mathcal{D}$ is split into a training set $\mathcal{D}_{\text{train}}$ for proposing skill updates, a validation set $\mathcal{D}_{\text{val}}$ for selecting skill updates, and a held-out test set $\mathcal{D}_{\text{test}}$ for final evaluation. 
$\mathcal{D}_{\text{train}}$ is further divided into non-overlapping batches $\{B_1, B_2, \ldots, B_T\}$. A different batch is used at each iteration to reveal different error patterns. Batch-based training can also reduce overfitting to individual responses and encourages general rules.

Across iterations, we maintain the best $\Delta$ seen so far, denoted $\Delta_{\text{best}}$, and the corresponding best skill $s_{\text{best}} = s_0 \oplus \Delta_{\text{best}}$.
The optimization begins with $\Delta_{\text{best}} = \emptyset$, so the initial best skill is simply $s_{\text{best}}=s_0$.
We first evaluate this initial skill on $\mathcal{D}_{\text{val}}$ to establish the reference QWK.
At each iteration $t$, the system executes the following steps (illustrated in Figure~\ref{architecture}).



\paragraph{Step 1: Rubric Generation.}
The current best skill $s_{\text{best}} = s_0 \oplus \Delta_{\text{best}}$ is used to generate a rubric for the item $q$:
\begin{equation}
    R_t = \text{LLM}_{\text{gen}}(s_{\text{best}},\; q) \, .
\end{equation}
 
\paragraph{Step 2: Batch Scoring.}
The generated rubric $R_t$ is used to score all responses in the current training batch $B_t$:
\begin{equation}
    \hat{y}_i = \text{LLM}_{\text{score}}(R_t,\; q,\; x_i), \quad \forall\, x_i \in B_t \, .
\end{equation}
For each response, the LLM scorer also produces a brief justification $j_i$ explaining its decision. These justifications are critical for downstream diagnosis, because they allow the diagnoser to distinguish whether a scoring error stems from the rubric or from how the scorer used it.
 
\paragraph{Step 3: Evaluation.}
Predicted scores are compared against human scores to produce error statistics, including overall accuracy, the distribution of over- and under-scoring, and confusion patterns between specific score levels (e.g., human $2 \to$ predicted $3$).
We denote the set of mis-scored response indices in iteration $t$ as $\mathcal{E}_t$, and collect the corresponding responses together with their justifications for diagnosis.

\paragraph{Step 4: Diagnosis and Skill Update.}
The error statistics from Step 3 describe what went wrong but not why.
The same confusion pattern may stem from different rubric flaws such as missing elements, ambiguous boundaries, or overly lenient wording, and each requires a different correction.

To identify actionable root causes, the diagnoser receives the error statistics $\mathcal{S}_t$, the mis-scored cases with their justifications $\{(x_i, \hat{y}_i, y_i, j_i)\}_{i \in \mathcal{E}_t}$, the current skill $s_{\text{best}}$, and the generated rubric $R_t$. It clusters the errors by underlying cause rather than surface-level symptoms, traces each cluster back to a specific gap or flaw in the current $\Delta$, and finally outputs an updated $\Delta_{t}^{\text{cand}}$:
\begin{equation}
    \Delta_{t}^{\text{cand}} = \text{LLM}_{\text{diag}}\big(s_{\text{best}},\; \mathcal{S}_t,\; R_t,\; \{(x_i,\; \hat{y}_i,\; y_i,\; j_i)\}_{i \in \mathcal{E}_t})
\end{equation}
Crucially, the diagnoser is instructed to target its modifications at the skill level, identifying which rubric construction rules to add, remove, or revise, rather than proposing item-specific rubric edits. The updated $\Delta_{t}^{\text{cand}}$ must remain content-free, that is, it may not reference any topic or detail from the current item, so that the learned rules remain transferable to new items.
 
\paragraph{Step 5: Validation Gatekeeper.}
Since the diagnoser's proposals are not guaranteed to improve scoring, a validation step is needed to prevent skill degradation.
The candidate skill $s_{cand}=s_0 \oplus \Delta_{t}^{\text{cand}}$ is used to generate a new rubric and score $\mathcal{D}_{\text{val}}$. Let $\text{QWK}(s)$ denote the QWK obtained by using skill $s$ to generate a rubric and score $\mathcal{D}_{\text{val}}$. The candidate is accepted if it exceeds the current best:
\begin{equation}
    \Delta_{\text{best}} \leftarrow
    \begin{cases}
        \Delta_t^{\text{cand}} & \text{if } \text{QWK}(s_{\text{cand}}) > \text{QWK}(s_{\text{best}}) \\
        \Delta_{\text{best}} & \text{otherwise}
    \end{cases}
\end{equation}
If accepted, $\Delta_{\text{best}}$ is updated and the failure counter is reset. If rejected, $\Delta_{\text{best}}$ is retained and the failure counter is incremented. The process terminates when all batches have been used or when validation performance has not improved for three consecutive rounds (early stopping).

\section{Experiments and Analysis}
Our experiments evaluate two questions.
First, we evaluate whether an LLM can learn a skill to construct effective rubrics for an assessment item from scoring practice with human-scored responses only, without requiring an expert-written rubric.
Here, the skill is learned and tested on the same item, measuring whether it can match or exceed an expert rubric (Sections~\ref{sec:main-results}--\ref{sec:robustness}).
Second, we evaluate whether the learned skill is item-independent and transferable to unseen items by applying a skill optimized on one item directly to other items without further optimization (Section~\ref{sec:transfer}).

\subsection{Experimental Setup}
\label{sec:setup}

\paragraph{Dataset.}
We evaluate our framework on the Automated Student Assessment Prize Short Answer Scoring (ASAP-SAS) dataset, which contains 17,043 student responses across 10 short-answer items. Each item has a discrete score range (0--2 or 0--3) and includes a dataset-provided expert rubric. The number of responses per item ranges from approximately 1,300 to 1,800. For each item, we split the data into training (65\%), validation (15\%), and test (20\%) sets using stratified sampling to preserve score distributions. The training set is further divided into non-overlapping batches of approximately 100 responses each.
 
\paragraph{Models.}
Our framework uses two LLMs in distinct roles. The \textit{scorer and generator} is Qwen3.5-9B~\cite{yang2025qwen3}, served locally via vLLM~\cite{kwon2023efficient} on two NVIDIA RTX A6000 GPUs (48GB each), with a context length of 32,768 tokens and greedy decoding (temperature=0). The \textit{diagnoser} is GPT-5.4 Thinking~\cite{openai2026gpt54thinking}, accessed through the OpenAI API. This asymmetric design is intentional. The scorer is a smaller, cost-efficient model representative of what practitioners would deploy at scale, while the diagnoser is a stronger model whose reasoning capability is needed only during the offline optimization phase.
 
\paragraph{Initial skill ($s_0$) variants.}
We experiment with three versions of the human-authored scaffold $s_0$, varying in the level of detail provided:
\begin{itemize}
    \item \textbf{Weak:} a single-sentence instruction.
    \item \textbf{Medium:} a brief paragraph identifying the key step.
    \item \textbf{Strong:} a detailed five-step procedure covering task analysis, key element identification, score level definition, descriptor writing, and variation anticipation.
\end{itemize}
All three variants are shared across all 10 items without modification. The full text of each variant is provided in Appendix~\ref{app:s0}, and all prompt templates are provided in Appendix~\ref{app:prompts}.
 
\paragraph{Scoring conditions.}
We compare four conditions: (1)~\textit{w/o rubric}: the scorer receives only the item and student response, with no rubric; (2)~$s_0$: the scorer uses a rubric generated by the initial skill before optimization; (3)~$s_{\text{best}}$: the scorer uses a rubric generated by the optimized skill at the end of optimization; and (4)~\textit{expert rubric}: the scorer uses the expert-crafted rubric provided with the dataset.
 
\paragraph{Evaluation.}
We report Quadratic Weighted Kappa (QWK) on the held-out test set as the primary metric, consistent with the optimization objective. For items 1, 7, and 9, we run the optimization with three random seeds (42, 63, 168) to assess stability. For the remaining items, we report results with seed 42. All reported test scores are computed using the final $s_{\text{best}}$.
 
\paragraph{Optimization details.}
The optimization uses early stopping with a patience of three consecutive rejections. In practice, most runs converge within 3--7 rounds. The diagnoser is prompted to follow a structured four-phase analysis (explore errors, cluster by cause, explain root causes, revise the skill) and is constrained to output only content-free rubric construction rules that do not reference any item-specific detail.

\subsection{Main Results}
\label{sec:main-results}

\begin{table*}[t]
\caption{Test QWK across 10 ASAP-SAS items. We compare four conditions: no rubric, initial skill ($s_0$), optimized skill ($s_{\text{best}}$), and expert rubric, for three $s_0$ variants (weak, medium, strong). \textbf{Bold}: best per item; \underline{underline}: second best.}
\label{tab:results}
\centering
\resizebox{\textwidth}{!}{%
\begin{tabular}{lcccccccc}
\toprule
\textbf{item} & \textbf{w/o rubric} & \textbf{weak $s_0$} & \textbf{weak $s_{\text{best}}$} & \textbf{medium $s_0$} & \textbf{medium $s_{\text{best}}$} & \textbf{strong $s_0$} & \textbf{strong $s_{\text{best}}$} & \textbf{w/ expert} \\
\midrule
item~1  & 0.698 & 0.761 & \textbf{0.793} & 0.677 & 0.779 & 0.742 & 0.775 & \underline{0.781} \\
item~2  & 0.431 & 0.574 & \underline{0.708} & 0.477 & \textbf{0.733} & 0.519 & 0.519 & 0.642 \\
item~3  & 0.397 & 0.392 & \underline{0.469} & 0.260 & \textbf{0.473} & 0.338 & 0.448 & 0.317 \\
item~4  & 0.433 & 0.348 & \textbf{0.598} & \underline{0.525} & 0.480 & 0.420 & 0.404 & 0.405 \\
item~5  & \textbf{0.690} & 0.421 & 0.670 & 0.400 & 0.643 & 0.512 & 0.663 & \underline{0.689} \\
item~6  & 0.676 & 0.703 & \textbf{0.769} & 0.705 & \underline{0.756} & 0.646 & 0.728 & 0.708 \\
item~7  & 0.143 & 0.417 & \textbf{0.469} & 0.330 & 0.407 & 0.368 & \underline{0.427} & 0.296 \\
item~8  & 0.416 & 0.377 & \textbf{0.573} & 0.339 & 0.548 & 0.359 & \underline{0.552} & 0.463 \\
item~9  & \underline{0.591} & 0.292 & 0.567 & 0.146 & \textbf{0.656} & 0.111 & 0.575 & 0.563 \\
item~10 & 0.427 & 0.260 & 0.578 & 0.293 & 0.483 & 0.315 & \textbf{0.648} & \underline{0.620} \\
\bottomrule
\end{tabular}%
}
\end{table*}

\begin{figure}[t]
\includegraphics[width=\textwidth]{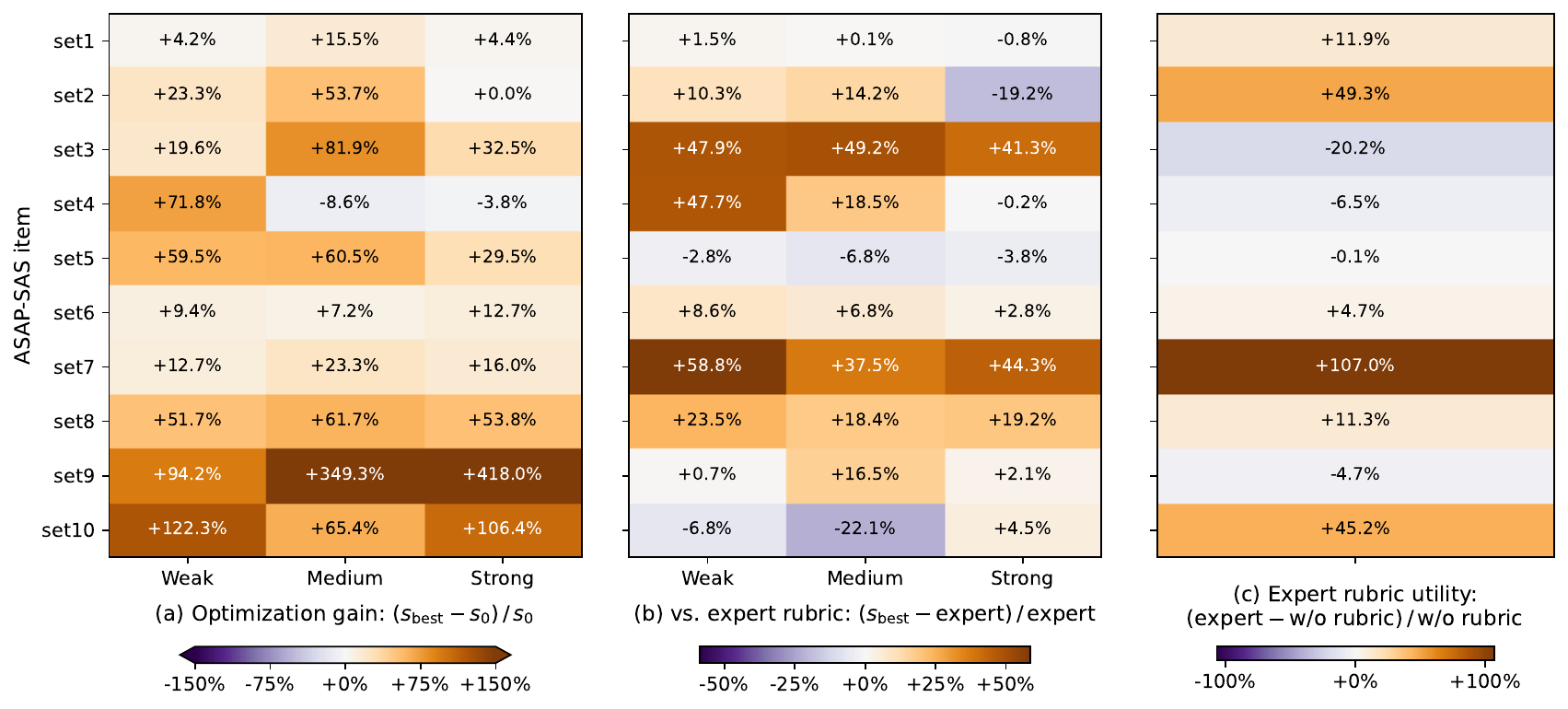}
\caption{
Relative performance changes (\%) across 10 ASAP-SAS items. (a) Optimization gain of $s_{\text{best}}$ over $s_0$ for each initial skill variant. (b) $s_{\text{best}}$ compared against the dataset-provided expert rubric. (c) Effect of the expert rubric relative to using no rubric. Relative changes are computed from the original unrounded QWK scores.
}
\label{fig:heatmap}
\end{figure}

\paragraph{Optimization consistently improves initial skills.}
Table~\ref{tab:results} reports test QWK across all 10 items under four conditions (w/o rubric, $s_0$, $s_{\text{best}}$, w/ expert). For each of the 10 items, we optimize skills starting from three different $s_0$ variants. This gives 30 optimized skills in total. Among them, 27 achieve higher QWK than their corresponding initial $s_0$, with a median relative gain of 31\% over $s_0$ (Figure~\ref{fig:heatmap}a).

\paragraph{Optimized skills frequently outperform expert rubrics for LLM scoring.}
 Across the 10 items, at least one optimized skill variant achieves the highest QWK in 9 items. The exception is item~5, where the no-rubric baseline is marginally higher.
 Figure~\ref{fig:heatmap}b directly compares each optimized skill $s_{\text{best}}$ with the expert rubric, showing that most $s_{\text{best}}$ achieve higher QWK than the expert rubric, with a median gain of 7.7\% over the expert rubric and particularly large gains on items~3, 4, and 7.
 These results suggest that the optimized skills produce rubrics that are better aligned with the LLM scorer.
 Figure~\ref{fig:heatmap}c reveals that the expert rubrics themselves are not always beneficial for LLM-based scoring. For items~3, 4, 5, and 9, using the expert rubric actually degrades performance compared to providing no rubric at all. Recent work has shown that LLM scorers achieve higher reliability with fine-grained, checklist-based rubrics than with holistic scoring criteria~\cite{tang2026designing}. The expert rubrics in ASAP-SAS were designed for human raters and rely on qualitative descriptors (e.g., ``a thoughtful and thorough examination of the text''). Our optimization process, by contrast, learns to produce rubrics with explicit element definitions and concrete boundary rules, which is a format better suited to LLM-based scoring.

\paragraph{Limitation: element-counting scaffolds on holistic items.}
The only two cases where optimization degrades performance (item~4 with medium $s_{\text{best}}$ and strong $s_{\text{best}}$) share a common cause. Item~4 asks scorers to judge the degree of a student's ``critical stance,'' which is a judgment about overall quality (holistic scoring).
However, both the medium $s_0$ and strong $s_0$ encode an element-identification workflow: they instruct the LLM to list key elements and assign scores based on how many are present.
Optimization tends to refine within this framework, tightening element definitions and sharpening boundaries, which effectively converts a holistic judgment task into an element-counting task.
The confusion matrices support this interpretation. For the item~4 run initialized with medium $s_0$, over-scoring errors (human~0 $\rightarrow$ predicted~1) drop from 51 to 9 after optimization, but under-scoring errors (human~1 $\rightarrow$ predicted~0) increase from 11 to 82. This suggests that the stricter element requirements reject responses that human raters credited for partial understanding. By contrast, weak $s_0$ provides no procedural constraint, leaving the optimization free to explore different scoring strategies, and it produces the best result on item~4.

 
\subsection{Robustness Analysis}
\label{sec:robustness}

 The main results in Sections~\ref{sec:main-results} report performance under a single data split per item. In this section, we examine whether those conclusions are stable under two sources of variation: changes in the data split (cross-seed stability) and stochasticity in the diagnoser (cross-run stability).

 
\paragraph{Cross-seed stability.}
We conduct this analysis on items~1, 7, and 9, which represent different patterns of optimization behavior: modest gains near expert-rubric performance (item~1), consistent gains over the expert rubric (item~7), and large gains from low starting points (item~9). Each item is optimized with three seeds (42, 63, 168), varying the train/validation/test split.
Across all three items, $s_{\text{best}}$ \textbf{consistently outperforms its corresponding $s_0$ regardless of seed}. On item~1, the mean $s_{\text{best}}$ QWK across seeds is 0.785 (weak), 0.787 (medium), and 0.779 (strong), all approaching the expert rubric mean of 0.792, with standard deviations between 0.012 and 0.020. On item~7, all $s_{\text{best}}$ variants exceed the expert rubric by a substantial margin (mean $s_{\text{best}}$: 0.430--0.457 vs.\ expert: 0.303). On item~9, the same pattern holds. Mean $s_{\text{best}}$ ranges from 0.578 to 0.606, compared to an expert rubric mean of 0.526.
 
A particularly notable pattern emerges in the convergence of $s_0$ variants after optimization. Before optimization, the spread across variants is large: on item~9, $s_0$ ranges from 0.086 (strong) to 0.321 (weak), a gap of 0.235. After optimization, $s_{\text{best}}$ ranges from 0.578 to 0.606, narrowing the gap to 0.028. Item~1 shows the same effect: $s_0$ spread of 0.090 narrows to 0.008 after optimization. This suggests that \textbf{the optimization can largely compensate for differences in initial skill quality}.
 
\paragraph{Cross-run stability.}
Since the diagnoser (GPT-5.4 Thinking) is a reasoning model for which temperature adjustments are not supported, and API outputs are not guaranteed to be fully deterministic, its outputs may vary across runs even with identical inputs. To quantify this effect, we run three independent optimizations on item~1 with seed~42. The standard deviation of $s_{\text{best}}$ across runs is 0.013 (weak), 0.009 (medium), and 0.022 (strong), comparable in magnitude to the cross-seed variation. This suggests that diagnoser stochasticity is not a larger source of variance than the data split itself. However, individual runs can occasionally produce weaker results. One strong-variant run yields $s_{\text{best}} = 0.728$, a slight degradation from $s_0 = 0.742$, while the other two runs both improve to 0.775. This reinforces that the optimization outcome, while generally positive, is not guaranteed in every single run.

\subsection{Cross-Item Transfer}
\label{sec:transfer}
 
We investigate whether the optimized $\Delta$ transfers across items by applying the $s_{\text{best}}$ learned on one item to all other items. Full results are provided in Appendix~\ref{app:transfer}.

\paragraph{Transferred skills often improve over unoptimized $s_0$ on the target item.}
 When a skill optimized on one item is applied to a different item, 62–76\% of such transfers show improvement over the target item's unoptimized $s_0$, with a median relative gain of 11--18\% depending on the $s_0$ variant.
 This suggests that $\Delta$ captures generally useful scoring rules, such as how to handle vague phrasings or when to merge overlapping elements, that remain effective on unseen items.

\paragraph{Transferability increases with stronger scaffolds.}
However, transferred skills rarely match the performance of directly optimizing on the target item. An interesting gradient emerges across $s_0$ variants. With weak $s_0$, only 3\% of transfers match or exceed the target's own $s_{\text{best}}$; with medium, 16\%; with strong, 27\%. This suggests a trade-off between in-distribution performance and transferability. Weak $s_0$ grants the optimization maximum freedom, enabling it to learn rules highly adapted to the source item. This is reflected in Table~\ref{tab:results}, where weak $s_{\text{best}}$ achieves the best or second-best QWK on 7 out of 10 items. However, these highly adapted rules are also the least portable. Strong $s_0$, by contrast, constrains $\Delta$ to fine-grained adjustments within a fixed framework, yielding more modest in-distribution gains but more transferable rules.
 

\section{Conclusion and Discussion}
\label{sec:conclusion}
 
We introduced the concept of assessment skills as reusable, natural-language-encoded procedural knowledge that guides LLMs in executing specific stages of the scoring workflow. Focusing on rubric construction as an initial case study, we proposed an iterative optimization framework that decomposes the skill into a fixed human-authored scaffold $s_0$ and a learnable refinement $\Delta$, and iteratively learns $\Delta$ through error diagnosis and validation gating.
Experiments on 10 ASAP-SAS items show that the optimized skill $s_{\text{best}}$ achieves the highest QWK in 9 out of 10 items across all conditions, frequently surpassing the dataset-provided expert rubric. This improvement stems not from producing objectively superior rubrics, but from generating rubrics better suited to LLM-based scoring. The optimized skills produce rubrics with explicit element definitions and concrete boundary rules that LLMs can execute more consistently than qualitative descriptors designed for human raters.

Several limitations point to directions for future work. First, the current $s_0$ is built around an element-identification workflow, which does not fit holistic rubric types that evaluate responses as a matter of degree rather than by counting discrete elements. Extending the framework to support multiple rubric paradigms, or allowing the optimization to select an appropriate paradigm automatically, would broaden applicability. Second, cross-item transfer experiments show that while transferred skills outperform unoptimized $s_0$ on new items in the majority of configurations, they fall short of direct optimization. The current setup evaluates only single-source transfer. Aggregating skills learned across many items to extract shared rules while filtering out item-specific patterns is a natural next step. Third, rubric construction is only one stage of the scoring workflow. Applying the same skill-optimization approach to other stages, such as evidence identification or feedback generation, remains unexplored.


\appendix
\section{Initial Skill Variants}
\label{app:s0}
 
We experiment with three versions of the human-authored scaffold $s_0$. Their full texts are provided below.
 
\begin{tcolorbox}[title=Weak $s_0$, colback=gray!5, colframe=gray!100, fonttitle=\bfseries\small, fontupper=\small]
Generate a scoring rubric for the test item.
\end{tcolorbox}
 
\begin{tcolorbox}[title=Medium $s_0$, colback=gray!5, colframe=gray!100, fonttitle=\bfseries\small, fontupper=\small]
You are an expert in educational assessment. Given a test item, generate a scoring rubric that can be used to grade student responses.
 
Identify the key pieces of evidence a correct response should contain, and map them to score levels from highest to lowest.
\end{tcolorbox}
 
\begin{tcolorbox}[title=Strong $s_0$, colback=gray!5, colframe=gray!100, fonttitle=\bfseries\small, fontupper=\small]
You are an expert in educational assessment. Given a test item, generate a scoring rubric that can be used to grade student responses.
 
Follow these steps:
 
1. ANALYZE THE TASK: Read the item carefully. Identify what the student is being asked to do (e.g., list, describe, explain, compare, analyze). Determine the subject area and the core concept being assessed.
 
2. IDENTIFY KEY ELEMENTS: Based on the task requirements and your domain knowledge, list all the specific, scorable pieces of evidence that a complete and correct response should contain. Each key element should be an independently verifiable claim or piece of information.
 
3. DEFINE SCORE LEVELS: Map the key elements to score levels. Use the number of key elements addressed as the primary basis for distinguishing score levels. Assign the highest score to responses that address all or nearly all key elements, and the lowest score to responses that address none.
 
4. WRITE SCORE DESCRIPTORS: For each score level, write a brief descriptor that specifies what a response at that level looks like, referencing the key elements.
 
5. ANTICIPATE VARIATION: List common alternative phrasings, partial understandings, or borderline cases that scorers may encounter. Specify how these should be handled.
 
OUTPUT FORMAT:\\
- Key Elements: [numbered list]\\
- Scoring Scale: [score levels with descriptors]\\
- Scoring Notes: [edge cases and acceptable variations]
\end{tcolorbox}
 
\section{Prompt Templates}
\label{app:prompts}
 
\begin{tcolorbox}[title=Rubric Generation Prompt, colback=blue!3, colframe=blue!100!black, fonttitle=\bfseries\small, fontupper=\small]
\texttt{\{skill\}}
 
Based on the following test item, generate a scoring rubric.
 
ITEM:\\
\texttt{\{question\}}
\end{tcolorbox}
 
\begin{tcolorbox}[title=Scoring Prompt, colback=blue!3, colframe=blue!100!black, fonttitle=\bfseries\small, fontupper=\small]
You are an expert grader for science open-ended items.
 
ITEM:\\
\texttt{\{question\}}
 
SCORING RUBRIC:\\
\texttt{\{rubric\}}
 
STUDENT RESPONSE:\\
\texttt{\{response\}}
 
Score this response on the scale defined in the rubric. Provide a brief justification (1--2 sentences), then on the last line output your final score --- a single digit --- enclosed in double square brackets, like [[X]] where X is the digit.
\end{tcolorbox}
 
\begin{tcolorbox}[title=Diagnosis and Skill Update Prompt, colback=blue!3, colframe=blue!100!black, fonttitle=\bfseries\small, fontupper=\small]
You are improving a general-purpose rubric generation skill by analyzing a scoring trial.
 
CURRENT SKILL:\\
\texttt{\{skill\}}
 
CURRENT RUBRIC (generated by this skill for one specific item):\\
\texttt{\{rubric\}}
 
ERROR STATISTICS:\\
\texttt{\{error\_stats\}}
 
ALL ERROR CASES:\\
\texttt{\{all\_errors\}}
 
You are a data analyst. Work through the errors in this order:

1. EXPLORE
   Read all error cases. Report what you observe about the distribution:
   \- Which confusion patterns are most common? (human score $\rightarrow$ model score)
   \- Which direction dominates (over-scoring, under-scoring, or mixed)?
   \- Are there structural features shared by many errors (response length,
     presence of specific reasoning patterns, types of content)?
   Quote specific cases to support each observation.

2. CLUSTER
   Group the errors into a small number (2–5) of error clusters, where
   each cluster contains errors that plausibly share the same underlying
   cause. Name each cluster by the pattern it represents, not by its
   symptom. List which specific cases belong to each cluster.

3. EXPLAIN EACH CLUSTER
   For each cluster, answer:
   - What is the underlying failure pattern that produces these errors?
   - What does the rubric do wrong in the face of this pattern?
   - What does the current skill fail to instruct that led the rubric
     to be this way?

4. REVISE THE SKILL
   For each cluster's root cause, propose a modification to the skill.
   The modification must:
   - Target a skill instruction, not a rubric content item
   - Be general (work for any subject, any item)
   - Be concrete enough that the next rubric generation would behave
     differently

   Output the full revised skill.

During steps 1–3 you may quote specific student responses, scores, and
reasoning. The final revised skill in step 4 must not mention any topic
or content from this specific item.
 
UPDATED AUGMENTATION:
\end{tcolorbox}

\section{Cross-Item Transfer Results}
\label{app:transfer}

\begin{figure*}[h]
\centering
\includegraphics[width=\textwidth]{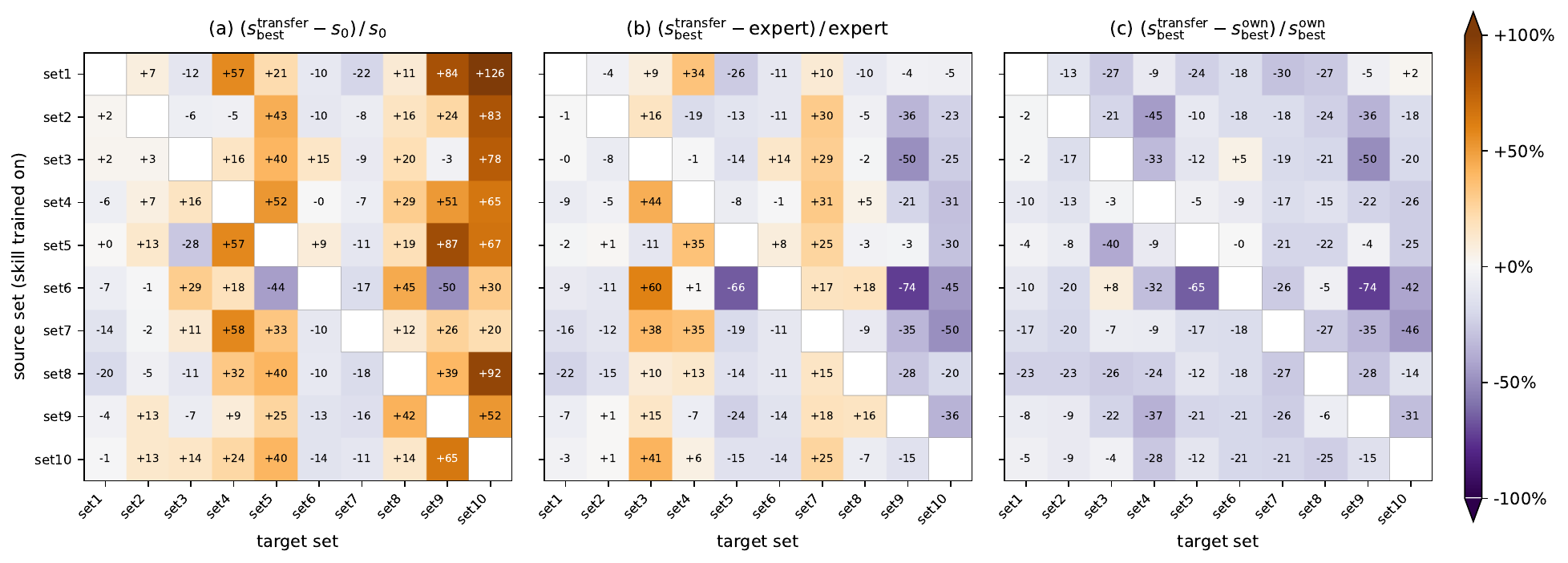}
\caption{Cross-item transfer results with weak $s_0$.}
\label{fig:transfer-weak}
\end{figure*}

\begin{figure*}[h]
\centering
\includegraphics[width=\textwidth]{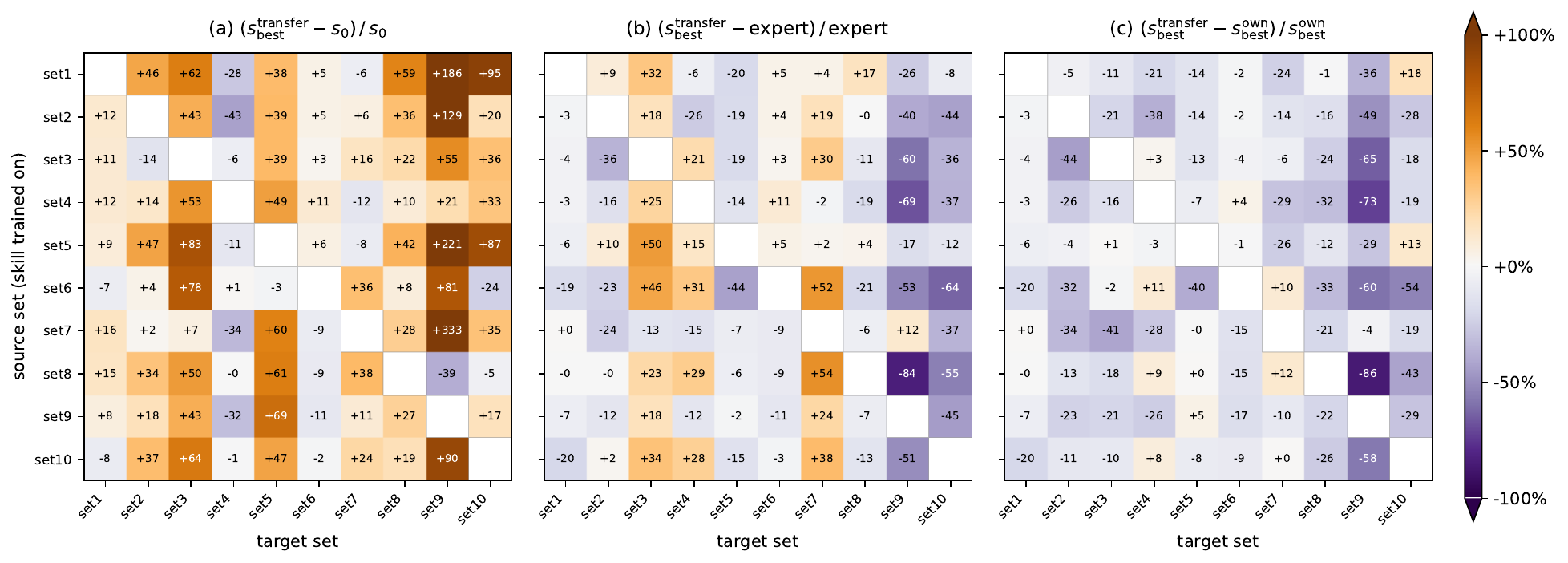}
\caption{Cross-item transfer results with medium $s_0$.}
\label{fig:transfer-medium}
\end{figure*}

\begin{figure*}[h]
\centering
\includegraphics[width=\textwidth]{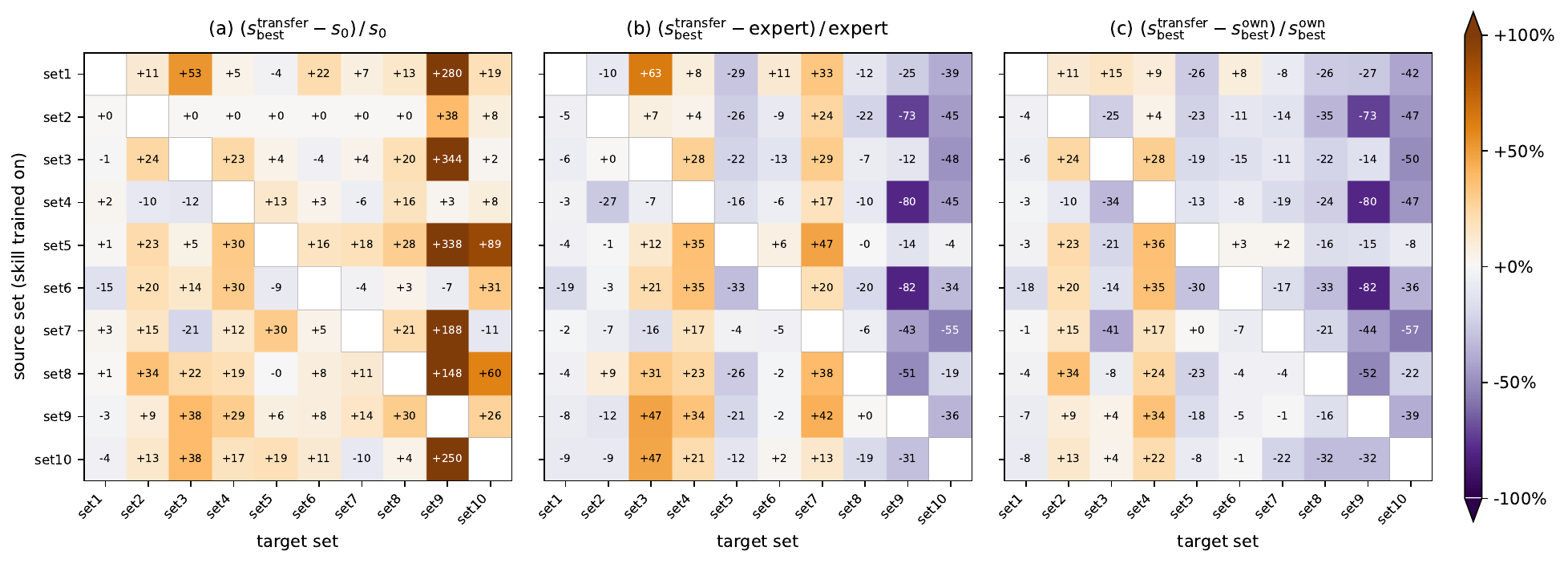}
\caption{Cross-item transfer results with strong $s_0$.}
\label{fig:transfer-strong}
\end{figure*}

Figures~\ref{fig:transfer-weak}, \ref{fig:transfer-medium}, and \ref{fig:transfer-strong} show cross-item transfer results for $s_{\text{best}}$ skills optimized from weak, medium, and strong $s_0$ respectively. In each heatmap, a cell at row $i$, column $j$ represents the result of applying the $s_{\text{best}}$ optimized on item $i$ to item $j$. Panel~(a) measures the relative gain over the target item's unoptimized $s_0$, panel~(b) over the expert rubric, and panel~(c) over the target item's own $s_{\text{best}}$. Diagonal entries (boxed) are not transferred and correspond to in-distribution optimization.

\end{document}